# Visual Question Answering (VQA) on Images with Superimposed Text

Venkat Kodali[1] and Daniel Berleant[2]

[1] Department of Information Science, University of Arkansas at Little Rock, USA
vxkodali@ualr.edu
[2] Department of Information Science, University of Arkansas at Little Rock, USA
jdberleant@ualr.edu

**Abstract.** Superimposed text annotations have been under-investigated, yet are ubiquitous, useful and important, especially in medical images. Medical images also highlight the challenges posed by low resolution, noise and superimposed textual meta-information. Therefor we probed the impact of superimposing text onto medical images on VQA. Our results revealed that this textual meta-information can be added without severely degrading key measures of VQA performance. Our findings are significant because they validate the practice of superimposing text on images, even for medical images subjected to the VQA task using AI techniques. The work helps advance understanding of VQA in general and, in particular, in the domain of healthcare and medicine.

## 1   Introduction

In medicine, images such as from radiology play a critical role in reaching diagnostic conclusions about the presence and progression of disease. Limitations in the images stemming from characteristics of the imaging technology and of the individual patient can present significant challenges to diagnostic conclusions. Notably, sharpness and contrast of medical images are often low, even when compared with inexpert and casual snapshots of everyday things (Fig. 1). These limitations must be considered in developing AI systems that assist experts in image interpretation, such as physicians who seek to provide accurate diagnoses and informed treatment decisions.

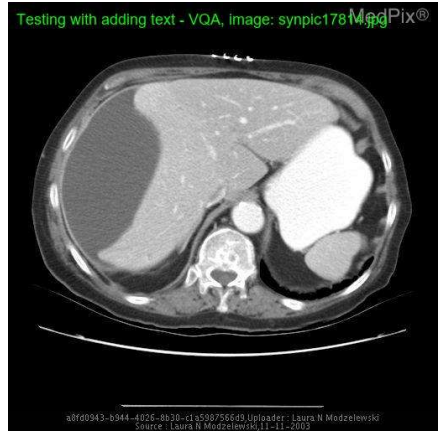

**Figure 1.** Example image with green added text.

VQA (visual question answering) is an image interpretation task that is increasingly recognized for its practical potential, as well as for being a touchstone task in AGI (artificial general intelligence) because it requires integrating visual processing and natural language. VQA's dramatic emergence is shown by Fig. 2. As that chart illustrates, the number of VQA papers appearing in 2022 increased by 23% compared to 2021, which itself showed a 20% increase over 2020. Indeed, the field has been expanding dramatically, starting shortly before 2015 [1][2][3]. That level of activity has produced significant progress in visual question answering architectures and their performance. In 2014, encoder decoder architecture was the state of the art for VQA models. In 2017, the field transitioned to transformers as the leading technology for VQA models [4]. In 2018, BERT was introduced, marking a significant advancement in the performance of VQA models. BERT replaced LSTM/RNN models in the field of natural language processing and emerged as the state-of-the-art approach for VQA models. In 2019, visual BERT was introduced, further enhancing the capabilities of VQA systems. In 2020, vision transformers were introduced, leveraging the BERT concept for image feature extraction. For this article, we have constructed BERT based VQA architecture for the experiments with superimposed text on images.

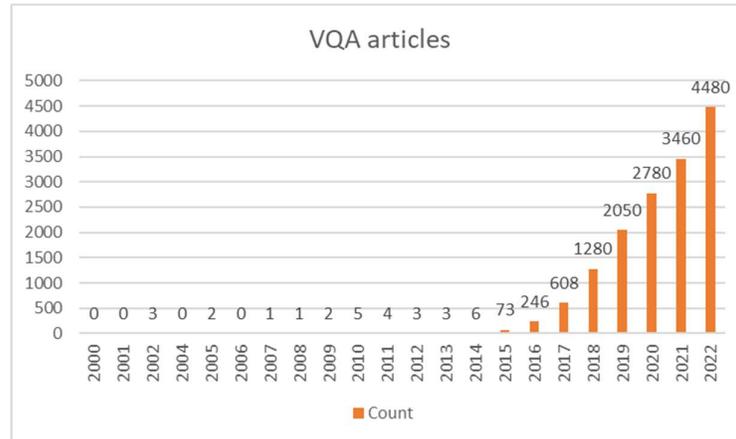

**Figure 2.** The expanding publication rate of VQA papers. Note that the histogram bars represent *new* articles each year (not the cumulative total of papers). This surely ranks VQA as one of the currently fastest growing fields. (Source: updated from Fig. 3 of Kodali & Berleant (2022) [5].)

To fulfill the role of an AGI touchstone problem, VQA needs to work on a wide variety of image types. Nevertheless, the amount of VQA research devoted to images with text is small compared to the VQA body of research as a whole.

Text that is naturally within images, such as images containing street signs, is one such image type that has attracted significant interest in recent years [6][7]. The problem has become known as textVQA. Furkan et al. (2019 [6]) observed that many (about 50%) of real images in datasets do in fact contain such text. Such text is likely to provide clues helpful in interpreting and answering question about the image.

A variant problem is images whose principal content is text (Mishra et al. 2019 [8]), a problem they call OCR-VQA. An example is book covers. The problem this body of work examines is how to use text that is an inherent part of an image to assist in understanding the image and thus answer questions about it (https://eval.ai/web/challenges/challenge-page/874/overview).

However, text present as an intrinsic part of an image is only one aspect of the overall problem of text in images. Text may instead be superimposed onto an image, corrupting the original, clean image. Such text might contain useful information that could improve the model's performance in answering questions about the image. However the added text might also present challenges to effective image analysis because the superimposed text may obscure image features or be confused with or misidentified as image features during the analysis process. Thus superimposing text is a form of image corruption. Therefor the problem of VQA on images containing text that is an inherent feature of the image (and thus cannot be viewed as corrupting) is distinct from the problem of VQA on images containing superimposed text annotations.

The problem of superimposed text is important due to the basic fact that many images, both medical (Fig. 1, etc.) and in general, do have added superimposed text annotations. These text annotations can serve various useful purposes, such as semantically meaningful annotations, patient identification codes, copyright notices, etc. Therefor better understanding of VQA on images with superimposed text annotations is needed. That task is the focus of the present article.

The next section provides further details on the problem of analyzing medical images with superimposed text. The section after that explains the underlying architecture used in the experiments. This is followed by sections on Experiment 1 (images without additional added text), Experiment 2 (text added to testing images but not training images), Experiment 3 (text added to both training and testing images) and Experiment 4 (text added to training but not testing images). The last section explains the conclusions.

## 2    Adding Superimposed Text onto Medical Images

The problem addressed in this article is analyzing medical images onto which text has been deliberately superimposed as human readable annotations. Medical images can be distorted due to various of reasons leading to corruptions associated with blur, contrast, hue, stretching [9], etc., Superimposed text annotations form one such corruption type. Common in medical images, these texts are not part of the actual image. Although they add information in the form of the content of the text annotations themselves, they overlay pixels in the original image, corrupting it and thus removing information from the image itself, risking reduced performance of visual question answering models.

To understand the impact on VQA of superimposing added text onto images, we conducted experiments to check. Fig. 3 shows a schematic of the VQA process that the present experiments used. In one experiment we added text to the training medical images and then tested on medical images without any added text. In another experiment, we trained our models using medical images without any added text, and then tested them on medical images that did have added text. An experiment in which both the training and test images had added text was also conducted. In these experiments, results were compared with the outcome of a control experiment in which no text was added to either the training set or the testing set.

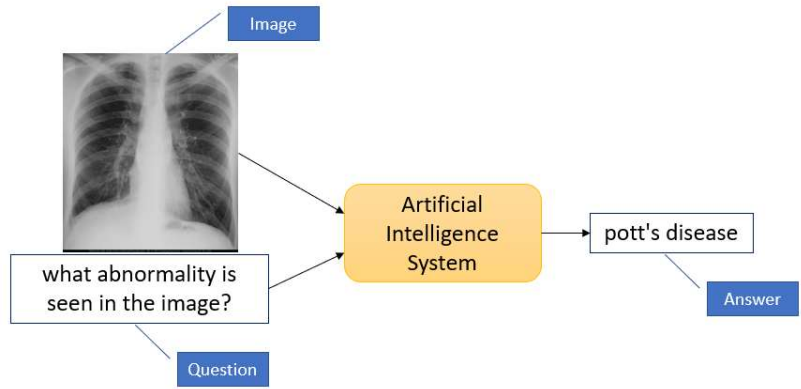
**Figure 3.** Visual question answering process.

Fig. 4 shows a sample training image prepared for use in the image analysis architecture by dividing it into patches.

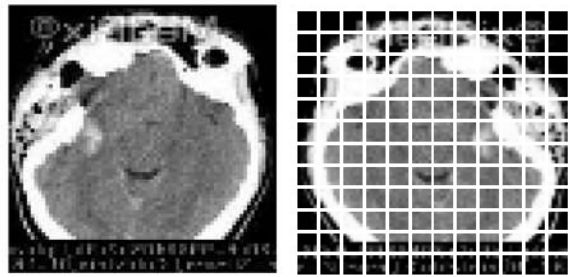
**Figure 4.** Image decomposed into patches.

## 3   Architecture of the model

The VQA architecture for this study was developed by combining components. Image features, input word IDs, input masks and input segment IDs form the inputs to the model. This was processed by a DenseNet-121, pretrained with weights derived from ImageNet. This was connected to a BERT based transformer model (Devlin et al. 2018 [10]), following up on previous work (Kodali 2022 [11]). Fig. 5 illustrates the transformer architecture. BERT models are of two types: BERT Base and BERT Large. BERT base uses 12 encoder layers and BERT large uses 24 encoder layers. BERT is pre-trained with two approaches: masked-language modeling to teach it relationships among words, and next sentence prediction to teach it about continuity from one sentence to the next. The experiments used the DenseNet-121 model (Huang et al. 2016 [12]), pretrained on ImageNet [13], to extract image features.

The optimizer parameter was specified as RMSProp. Finally, the VQA model was constructed by passing all the input vectors and output vectors from the DenseNet-121 component to the tensor flow BERT transformer model. Fig. 6 shows the visual question answering architecture used in the experiments.

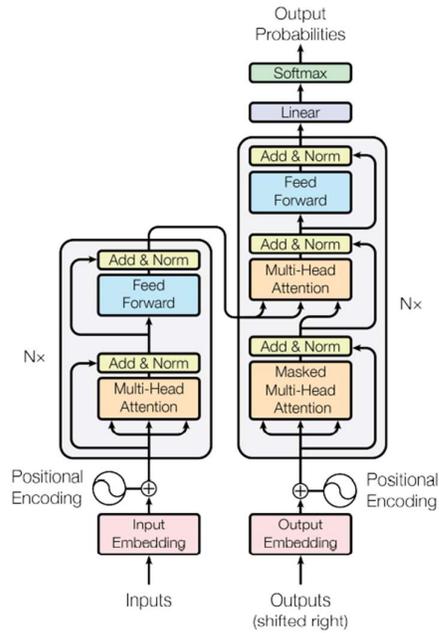

**Figure 5.** Transformer architecture (source: Vaswani et al. 2017 [4])

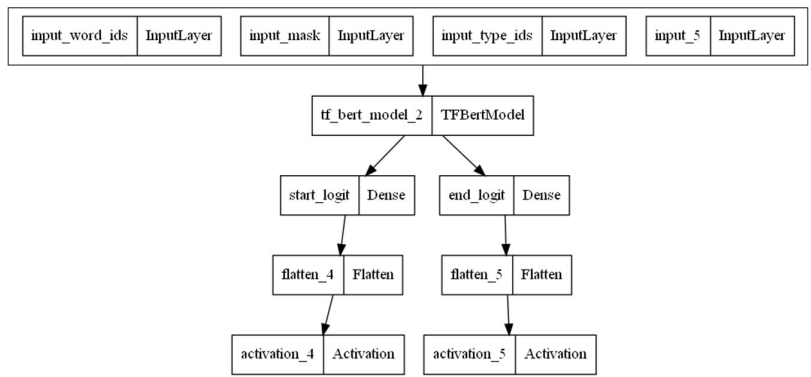

**Figure 6.** VQA architecture using BERT.

## 4 Experiment 1: Images Without Added Text

A control experiment was conducted with both training and testing on medical images without adding any new text to them. This provided baseline results to compare with results from other experiments in which the training images, test images, or both contained superimposed text added to the preexisting images. The medical image source was the 2019 ImageCLEF medical image set [14][15]. Hyperparameter values used in this experiment were batch size=32, epochs=100, optimizer=Adam, and learning rate=0.001. These hyperparameters values were also used in the other experiments reported in this article. The BLEU scores obtained from this experiment were used as baseline values for comparison with experiments on the later conditions in which text is superimposed onto the images. This enabled the impact of these text annotations to be assessed with respect to VQA performance.

From this experiment we obtained two accuracy [16][17] graphs, one for training and one for validation. Fig. 7 shows that training accuracy leveled off at close to 80%. Similarly, validation (i.e., testing) accuracy leveled off at 76%. Training and validation loss [18] started relatively high in the early iterations and rapidly decreased before leveling off.

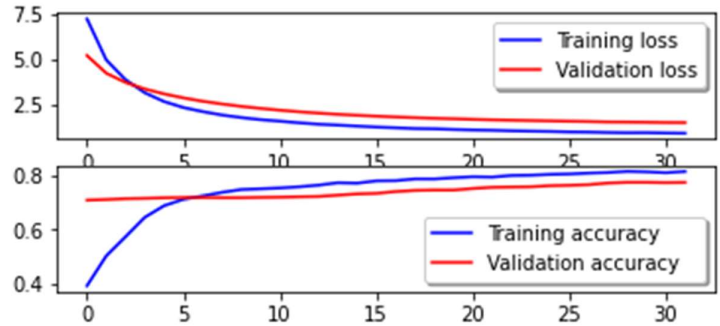

**Figure 7.** Loss and accuracy in control Experiment 1.

Fig. 7 (top) shows that both training loss and validation loss were high in the early iterations but rapidly decreased.

Table 1 shows the BLEU scores for Experiment 1. These BLEU scores provide baseline values useful for comparison with the subsequent experiments. A BLEU score of 0.56, for example, indicates that when tested on the VQA task, 56% of the time the system gave the correct answer [19].

In the 1-gram condition, sentence comparison is based on the 1-grams (1-word sequences) in the sentences. In the 2-gram condition, bigrams are used for comparison. In the 3-gram condition trigrams are used. The remaining condition we used specified the n-gram weights as (0.25, 0.25, 0.25, 0.25) which means comparing using 1-grams, 2-grams, 3-grams, and 4-grams, then taking a weighted average with equal (25% each) weights.

**Table 1.** BLEU scores for Experiment 1.

| Weight vectors | Conditions | BLEU Score |
|---|---|---|
| (1, 0, 0, 0) | 1-gram | 0.56 |
| (0, 1, 0, 0) | 2-grams | 0.28 |
| (0, 0, 1, 0) | 3-grams | 0.16 |
| (0.25, 0.25, 0.25, 0.25) | 1, 2, 3 & 4-grams | 0.21 |

## 5 Experiment 2: Testing on Images Containing Added Text

In this experiment, text was superimposed onto the medical images. The text was part of each test image, while the training images were clean without text annotations. This design models the situation in which medical images to be analyzed are annotated with text (such as patient identification, date, or other specific data) while the system is pretrained on a generic training set without patient-specific labeling. The risk is that superimposed text might interfere with the graphical information content of an image by acting as a kind of corruption or reduction in quality of the image [20], adversely affecting its analysis.

Fig. 8 shows graphs for training accuracy and loss as well as validation accuracy and loss. Fig. 8 (bottom) shows training accuracy rising quickly and then slowing as it reaches about 77%. Validation accuracy reaches about 75%.

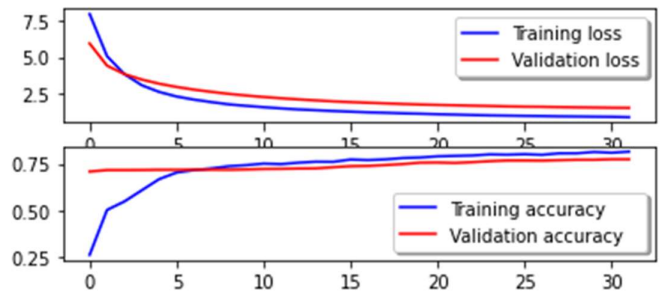

**Figure 8.** Loss and accuracy in Experiment 2.

Table 2 shows the BLEU scores for Experiment 2. BLEU scores for all conditions remained the same as for the baseline experiment shown in Table 1. Thus, in this experiment performance was not significantly impacted by the superposition of text annotations in the testing images.

Table 2. BLEU scores for Experiment 2.

| Weight vectors | Conditions | BLEU Score |
|---|---|---|
| (1, 0, 0, 0) | 1-gram | 0.56 |
| (0, 1, 0, 0) | 2-grams | 0.28 |
| (0, 0, 1, 0) | 3-grams | 0.16 |
| (0.25, 0.25, 0.25, 0.25) | 1, 2, 3 & 4-grams | 0.21 |

## 6      Experiment 3: Training and Testing on Images Containing Added Text

In this experiment text was superimposed onto medical images in both the training and the testing sets. This experimental design models the situation in which medical images to be analyzed are annotated with text as in the previous experiment), while the training set also has text annotations as might be done to make the training set more similar in its graphical properties to the images on which it is ultimately intended to be used.

The risk in this situation is parallel to the previous experiment in that superimposed text might act as a type of image corruption. In this framework, however, both the training and test sets are corrupted, potentially degrading performance even more but also potentially benefiting performance due to the increased similarity between the training and test sets.

Fig. 9 shows graphs for training accuracy and loss as well as validation accuracy and loss. Training accuracy (bottom) levels off at 80%, while validation accuracy reaches approximately 77%.

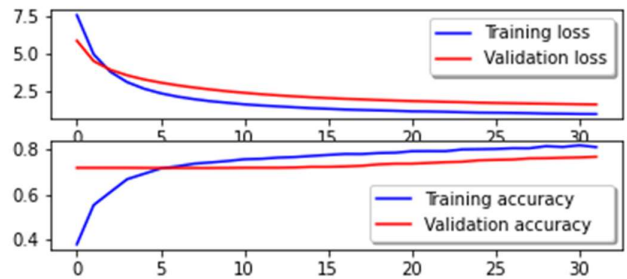

Figure 9. Loss and accuracy in Experiment 3.

Table 3 shows the BLEU scores for Experiment 3. In Table 3 the BLEU scores are slightly reduced for the 2-gram and 3-gram conditions compared to both the baseline experiment values shown in Table 1, and the test set with added text condition shown in Table 2. Thus, adding text to both training and testing medical images had a small impact on the performance of the visual question answering model.

**Table 3.** BLEU scores for Experiment 3.

| Weight vectors | Conditions | BLEU Score |
|---|---|---|
| (1, 0, 0, 0) | 1-gram | 0.56 |
| (0, 1, 0, 0) | 2-grams | 0.27 |
| (0, 0, 1, 0) | 3-grams | 0.15 |
| (0.25, 0.25, 0.25, 0.25) | 1, 2, 3 & 4-grams | 0.21 |

## 7    Experiment 4: Training on Images Containing Added Text

This experiment was conducted using a training set containing medical images onto which text was superimposed, but a test set of clean images without superimposed text. This experimental design models the situation in which a system is trained on images with added text in the expectation that it will be used later for analyzing images with added text, but then is used to analyze clean images instead. That could happen, for example, if there was a need to process images whose descriptions are stored separately instead of superimposed on the images.

The risk in this type of analysis scenario is that training on corrupted images will poison the learned neural network weights in that the ability to analyze clean images will be degraded. On the other hand, another possible outcome would be that the network would perform better on better images despite the training, much as a physician might benefit from improved imaging technologies despite having been trained in years past on images obtained from older imaging technologies.

Fig. 10 shows graphs for training accuracy and loss as well as validation accuracy and loss. Training accuracy (bottom)

leveled off at about 81%. Validation accuracy reaches about 71% by the end of the experiment.

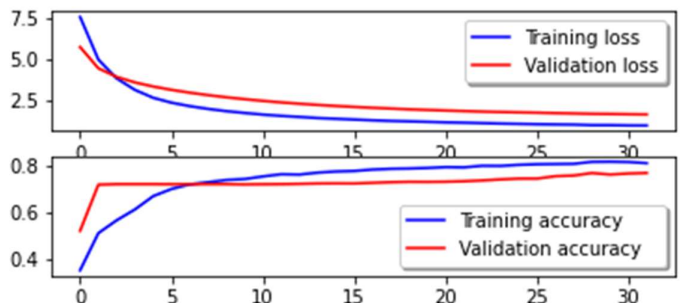
**Figure 10.** Loss and accuracy in Experiment 4.

Table 4 shows the BLEU scores for this experiment with text annotations added to training images. The table shows BLEU scores slightly reduced for 2-grams and 3-grams conditions compared to the baseline experiment values shown in Table 1, but on a par with Table 3 (in which text added to both the training and testing sets).

**Table 4.** BLEU scores for Experiment 4.

| Weight vectors | Conditions | BLEU Score |
|---|---|---|
| (1, 0, 0, 0) | 1-gram | 0.56 |
| (0, 1, 0, 0) | 2-grams | 0.26 |
| (0, 0, 1, 0) | 3-grams | 0.14 |
| (0.25, 0.25, 0.25, 0.25) | 1, 2, 3 & 4-grams | 0.2 |

## 8   Conclusions

This article focused on a specific problem: visual question answering (VQA) on images with superimposed text. When text was added to the test set, but training was on images without added text, there was no change in the BLEU scores. However, when text was added to both the training and test sets, there was a measurable, albeit small, tendency for the BLEU scores to decrease. Another experiment involved adding text only to the training set and testing on images without added text. This led to a fairly small decrease in VQA performance.

Overall, different experimental conditions had effects ranging from negligible to rather modest. Thus we have shown that it is possible to superimpose text annotations onto images without seriously affecting the ability of AI algorithms to analyze them. These findings suggest that superimposing text on medical images can provide useful information with minimal degradation in image analysis algorithm performance. However, overloading an image with a large amount of text, heavier fonts, or text that overlaps and obscures important content in the image would have to lead to more serious negative impacts on VQA performance. Further work is needed to identify the acceptable parameters for text annotation that govern how to do it safely. Another direction for future work is leveraging the information content of superimposed text rather than looking at only at its tendency to corrupt the image. This topic has important practical implications, as VQA and other advanced, AI-based analyses are expected to increasingly find their way into practical applications.

Although future advances in VQA will likely lead to new and higher performing models the present results are expected to have enduring relevance: The focus here is not on VQA performance itself, but rather at the meta level of performance *robustness* [20]. In the present article the performance robustness that was investigated is the impact of superimposed text annotations on VQA. Understanding this impact will be of continuing concern for VQA algorithms both present and future.

## Acknowledgment

Publication of this work was supported by the National Science Foundation under Award No. OIA-1946391. The content reflects the views of the authors and not necessarily the NSF.